\def\year{2021}\relax
\pdfoutput=1

\documentclass[letterpaper]{article}

\usepackage{aaai21}  
\usepackage{times}  
\usepackage{helvet} 
\usepackage{courier}  
\usepackage[hyphens]{url}  
\usepackage{graphicx} 
\usepackage{natbib}  
\usepackage{caption} 

\usepackage{xcolor}

\author{
    Haekyu Park\textsuperscript{\rm 1},
    Zijie J. Wang\textsuperscript{\rm 1},
    Nilaksh Das\textsuperscript{\rm 1},
    Anindya S. Paul\textsuperscript{\rm 2},\\
    Pruthvi Perumalla\textsuperscript{\rm 1},
    Zhiyan Zhou\textsuperscript{\rm 1},
    Duen Horng Chau\textsuperscript{\rm 1}\\
}

\affiliations{
    \textsuperscript{\rm 1}Georgia Institute of Technology,\\
    \textsuperscript{\rm 2}Intel\\
    \{haekyu, jayw, nilakshdas, pp329, zzhou406, polo\}@gatech.edu, \\
    anindya.s.paul@intel.com
}

\newcommand{\modeltitle}[0]{SkeletonVis}
\newcommand{\model}[0]{\textsc{\modeltitle}}

\newcommand{\skeletonview}[0]{\textit{Skeleton View}}
\newcommand{\monitorview}[0]{\textit{Timeline View}}
\newcommand{\overlapview}[0]{\textit{Comparison View}}
\newcommand{\separateview}[0]{\textit{Split View}}

\definecolor{green}{RGB}{50,149,54}
\definecolor{orange}{RGB}{255,143,40}
\definecolor{red}{RGB}{198,50,42}
\definecolor{agreen}{RGB}{74, 198, 148}
\definecolor{purple}{RGB}{158, 62, 177}
\definecolor{darkpurple}{RGB}{170, 70, 210}
\definecolor{aqua}{RGB}{87, 180, 181}
\definecolor{lightblue}{RGB}{67, 130, 181}
\definecolor{aqua}{RGB}{87, 180, 181}
\definecolor{lightblue}{RGB}{72, 123, 232}
\definecolor{hotpink}{RGB}{255, 83, 115}
\definecolor{teal}{RGB}{90, 200, 250}

\newcommand{\todo}[1]{\textcolor{red}{[#1]}}
\newcommand{\polo}[1]{\textcolor{red}{[#1 -Polo]}}
\newcommand{\haekyu}[1]{\textcolor{purple}{[#1 -Haekyu]}}
\newcommand{\jay}[1]{\textcolor{hotpink}{[#1 -Jay]}}
\newcommand{\nilaksh}[1]{\textcolor{purple}{[#1 -Nilaksh]}}

\newcommand{\mkclean}{
  \renewcommand{\haekyu}[1]{}
  \renewcommand{\jay}[1]{}
  \renewcommand{\nilaksh}[1]{}
  \renewcommand{\polo}[1]{}
  \renewcommand{\todo}[1]{}
}
\mkclean{}

\title{\modeltitle{}: Interactive Visualization for Understanding \\Adversarial Attacks on Human Action Recognition Models}

\begin{document}

\maketitle

\begin{abstract}
Skeleton-based human action recognition technologies are increasingly used
in video based applications, 
such  as  home  robotics, healthcare on aging population, and surveillance.
However, such models
are vulnerable to adversarial attacks, raising serious concerns for their use in safety-critical applications.
To develop an effective defense against attacks, it is essential to understand how such attacks mislead the pose detection models into making incorrect predictions.
We present \model{}, the first interactive system that visualizes how the attacks work on the models to enhance human understanding of attacks.
\end{abstract}
\section{Introduction}
\begin{figure*}[t]
    \begin{center}
        \includegraphics[width=7in]{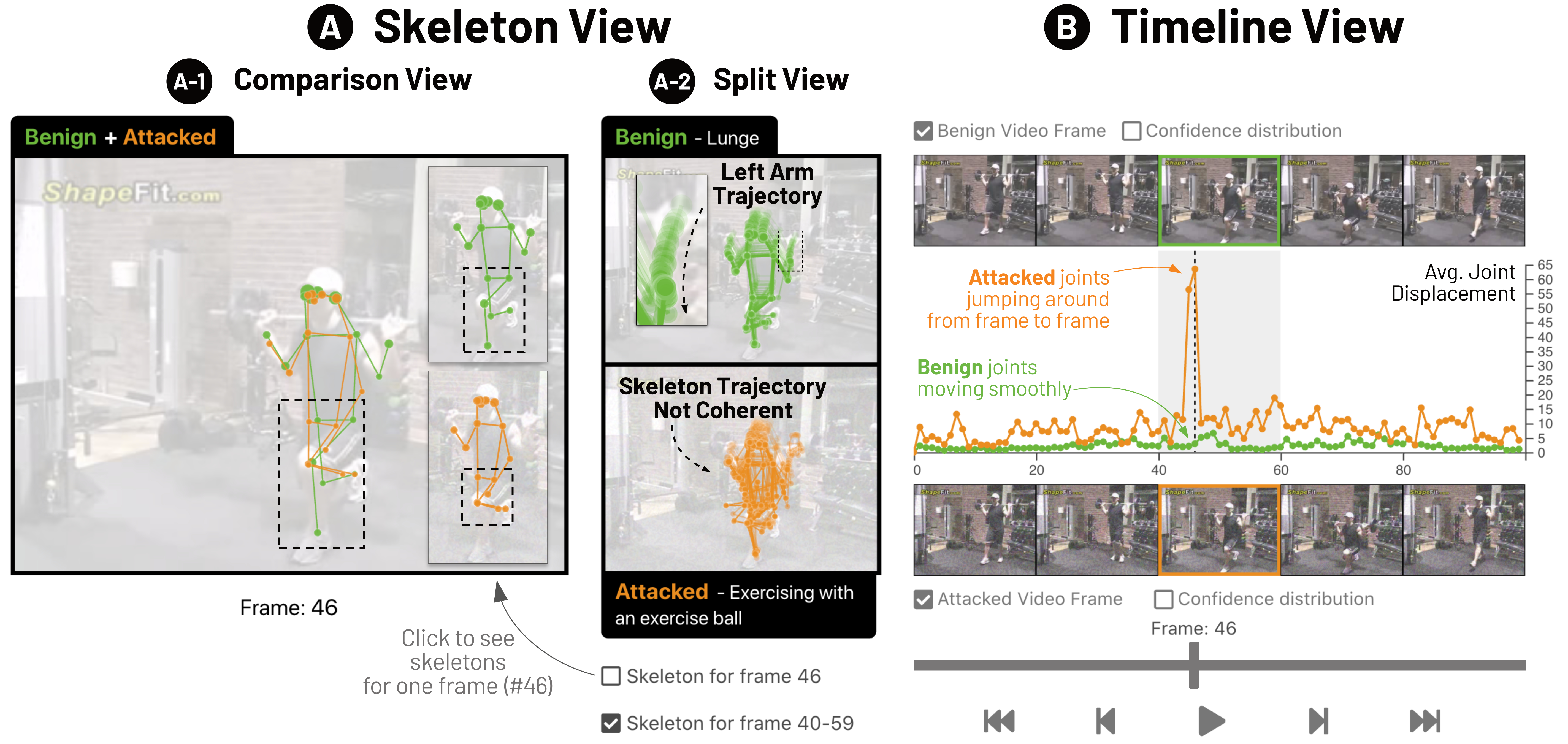}
        \caption{
        The interface of \model{}, visualizing how the \textit{Fast Gradient Method} 
        manipulates the left foot joints detected by the Detectron2 Keypoint R-CNN model. 
        (A) The \skeletonview{}  shows the joints perturbed to unexpected locations.
        (B) \monitorview{} reveals the attacked joints 
        spuriously jumping around from one frame to the next, 
        leading to a ``spike'' in the average joint displacement across attacked frames.
        These manipulations finally sway the ST-GCN action detection model 
        into misclassifying the attacked frames as ``exercising with exercise ball,''
        instead of the correct ``lunge'' classification.
        }
        \label{fig:teaser}
    \end{center}
\end{figure*}

Skeleton-based human action recognition technologies have been widely used in
many video understanding based applications, 
such as home robotics, eldercare, and surveillance
~\cite{yan2018spatial,choutas2018potion, saggese2019learning}.
These models have demonstrated promising prediction capabilities by analyzing the joints of human bodies
~\cite{yan2018spatial}. 
However, 
such techniques are vulnerable to adversarial attacks;
an attacker can add visually imperceptible perturbations to an input
and fool the models into arriving at incorrect predictions~\cite{liu2019adversarial, freitas2020unmask}.
This jeopardizes the use of human action recognition technologies in safety-critical applications, such as autonomous driving.
To make the models more robust against attacks, it is essential to understand how attacks inflict harm on the models.
Interactive visual analytics systems have great potential in helping users gain insights into adversarial attacks on deep learning models \cite{dasBluffInteractivelyDeciphering2020, das2020massif, park2019neuraldivergence}.
We present \model{} (Figure \ref{fig:teaser}), the first interactive visualization tool for understanding how adversarial attacks manipulate the input to
induce incorrect prediction.

\section{Demonstrating \model}

\model{}  takes as inputs a benign video clip (e.g., \textit{person lunging}) and its ``attacked'' version (e.g., \textit{exercising with a gym ball}) generated by an attack through perturbing pixels in benign clips (Figure \ref{fig:teaser}A). 
\model{} visualizes the detected human joints in the benign and attacked videos, and highlight how they may misalign and contribute to wrong predictions.
Furthermore, \model{} provides quantitative measurements across video frames to quantify the abnormal signals introduced in the attacked frames, 
which helps users more easily hone in on the specific frames exploited by the attacks (Figure \ref{fig:teaser}B).

\subsection{Extracting Human Joints to Predict Action}
\model{} supports end-to-end skeleton-based classification:
\begin{itemize}
    \item It first detects 17 human joints in each video frame, 
    using Detectron2's body joint R-CNN model ~\cite{wu2019detectron2}, 
    providing a spatial confidence map for each joint.
    \item It then combines the resulting spatial joint information from all video frames to infer the human action performed using ST-GCN~\cite{yan2018spatial}, a state-of-the-art model for skeleton-based action detection. 
\end{itemize}
Adversarial attacks are performed on the combined end-to-end model.

\subsection{\skeletonview{}: Explaining How Attacks Manipulate Detection of Human Joints}
\skeletonview{} (Figure \ref{fig:teaser}A) visually explains how the attacks are 
manipulating the human joints to cause misclassification, via a
\overlapview{} (Figure \ref{fig:teaser}A-1) 
and a
\separateview{} (Figure \ref{fig:teaser}A-2).

\overlapview{} enables \textit{spatial} comparison 
of how the detected joints are located differently in benign and adversarial frames by overlapping the two sets of joints.
For example, 
by revealing that the left foot is detected in unexpected locations in adversarial frames,
users can more easily glean insights into
why the pose detection model 
misclassifies the video 
(e.g., person lunging)
as an incorrect action (e.g., exercising with an exercise ball).

\separateview{} compares \textit{movement} of the human joints in the benign and adversarial clips.
For a clip, we visualize the trajectory of a human joint as follows.
Let $\mathbf{x}_1, \mathbf{x}_2, ... , \mathbf{x}_T$ be the sequence of the coordinates of a human joint in a clip.
We display the movement as in Figure \ref{fig:teaser}A-2, 
where joints from earlier frames (i.e., $\mathbf{x}_1, \mathbf{x}_2, ...$)
are visualized as more transparent dots,
and 
those from frames (i.e., $\mathbf{x}_{T}, \mathbf{x}_{T-1}, ...$)
are visualized as more opaque dots.
The joint trajectories and frames of the benign video are shown at the top, and those of the adversarial videos are shown at the bottom.

\subsection{\monitorview{}: Visualizing Abnormal Signals from Adversarial Attacks}
To help users discover attacked frames, 
\model{} quantifies and visualizes the pose detection models' responses to benign and attacked videos in the \monitorview{} (Figure \ref{fig:teaser}B),
enabling users to 
interactively compare 
and more easily pinpoint frames that they need to be wary of.

At the top of \monitorview{},
users can examine a \textit{video segment} (i.e., a range of frames) in detail, 
by clicking on the segment's thumbnail.
This highlights the thumbnail and its corresponding average joint displacement values in a line chart,
where the horizontal axis represents frames, and the vertical axis represents the displacement
from one frame to the next.

At the bottom of \monitorview{},
users can control which frame or segment to analyze through familiar controls used by typical video players.
The interactive timeline consists of a \textit{frame slider} and \textit{time buttons}.
Moving the knob on the \textit{frame slider} sets the selected frame in \overlapview{} (Figure \ref{fig:teaser}A-1)
and the segment in \separateview{} (Figure \ref{fig:teaser}A-2).
Pressing the ``Play'' button pauses or resumes the video playback.
The ``Fast backward'' and ``Fast forward'' buttons at both ends bring users to the very beginning and the very end of the clip respectively.
The ``Forward'' and ``Backward'' buttons step forward or backward by one frame.

\section{Engaging the Audience}
Our demonstration will focus on 
how even a basic adversarial attack, the Fast Gradient Method \cite{goodfellow2014explaining}, 
is able to manipulate a Detectron2 Keypoint R-CNN model 
into making incorrect predictions of human joint positions. 
As described earlier, the attack will be crafted by combining the human joint detection model 
with the ST-GCN skeleton-based action detection model.
\model{} will help the users better understand 
how the attack fools the downstream ST-GCN model 
into incorrectly classifying the action being performed, 
by directly perturbing 
pixels on an example video
from the UCF-101 action classification dataset \cite{soomro2012ucf101}.
For example, the audience of our demonstration will witness 
how the position of the detected feet is manipulated by the attack
by very slightly perturbing the video pixels,
that finally leads the model to misclassify 
``lunge'' action as ``exercising with gym ball'' (Figure \ref{fig:teaser}).

\section{Acknowledgments}
This work was supported in part by Defense Advanced Research Projects Agency (DARPA). Use, duplication, or disclosure is subject to the restrictions as stated in Agreement number HR00112030001 between the Government and the Performer.
This work was supported in part by NSF grants IIS-1563816, CNS-1704701, and gifts from Intel, NVIDIA, Google, Symantec, eBay, Amazon.

\bibliography{ref}
\end{document}